\documentclass[letterpaper]{article} 
\usepackage{aaai2027}  
\usepackage[hyphens]{url}  
\usepackage{graphicx} 
\usepackage{natbib}  
\usepackage{caption} 
\usepackage{algorithm}
\usepackage{algorithmic}

\usepackage{newfloat}
\usepackage{multirow} 
\usepackage{amsmath}
\usepackage{listings}
\DeclareCaptionStyle{ruled}{labelfont=normalfont,labelsep=colon,strut=off} 
\floatstyle{ruled}
\newfloat{listing}{tb}{lst}{}
\floatname{listing}{Listing}

\usepackage{booktabs}
\usepackage{algorithm}
\usepackage{algorithmic}
\usepackage[table]{xcolor}
\usepackage{amssymb}
\definecolor{gray94}{gray}{0.94}

\nocopyright 

\newcommand{\method}{\textsc{FailForge}}
\title{\method{}: Distilling Procedural Competence from Persistent Failures into Code Agents}

\author {
    Dongyi Lv\textsuperscript{\rm 1,\rm 2},
    Fushun E\textsuperscript{\rm 2},
    Aichen Cai\textsuperscript{\rm 2},
    Liang Huang\textsuperscript{\rm 2},
    Ya Zhang\textsuperscript{\rm 2},
    Qiuyu Ding,
    Canhui Wu\textsuperscript{\rm 1},\\
    Zhi Wang\textsuperscript{\rm 1},
    Yuesong Zhang\textsuperscript{\rm 2},
    Jiaqi Wang\textsuperscript{\rm 2},
    Nan Duan\textsuperscript{\rm 2}
}
\affiliations {
    \textsuperscript{\rm 1}Xi'an Jiaotong University,\ \ \ 
    \textsuperscript{\rm 2}Joy Future Academy, JD\\
    lvdongyi@stu.xjtu.edu.cn
}

\begin{document}

\maketitle
\begin{abstract}
Rejection sampling fine-tuning (RFT) is widely used to train code agents by generating trajectories on verifiable software engineering tasks, retaining those that pass the tests, and fine-tuning on the successful rollouts. However, even strong code agents repeatedly fail on a substantial fraction of such tasks, and standard RFT simply discards these failures. The discarded samples are precisely the hardest and most informative ones, drawn from verifiable instances that are costly to curate. Stronger base models may reduce the number of failures, but the remaining hard cases still define the frontier for further improvement. We propose \method, an agentic framework that converts failed rollouts into training signal. For each failed instance, an agent diagnoses the failure from error feedback and execution traces, distills the diagnosis into a concise and actionable skill, and injects the skill into the agent context for a guided second attempt. Trajectories that succeed under skill guidance are folded back into the RFT corpus. Crucially, the skill is removed at training time, so the model internalizes the recovered behavior rather than relying on external hints at inference. \method\ recovers over 26\% of previously failed instances at marginal additional cost, and training Qwen3.5-4B on the augmented corpus improves the SWE-bench Verified resolve rate by 6.6 points over a strong RFT baseline, with gains concentrated on the hardest problems.
\end{abstract}

\section{Introduction}
Large language model (LLM) agents have made rapid progress on real-world software engineering (SWE) tasks~\cite{chen2021codex,zhang2024codeagent,huang2023agentcoder}. These agents can autonomously navigate repositories, edit code, and validate fixes against test suites. Training such agents increasingly relies on rejection fine-tuning (RFT)~\cite{team2025kimi,zeng2026glm}. For each task, the model samples multiple trajectories in an executable environment. Trajectories that pass verification, such as the unit tests of a task, are retained as training data, while the rest are mostly discarded. Execution-based verification yields reliably correct supervision at scale without human annotation.
\begin{figure}
    \centering
    \includegraphics[width=\linewidth]{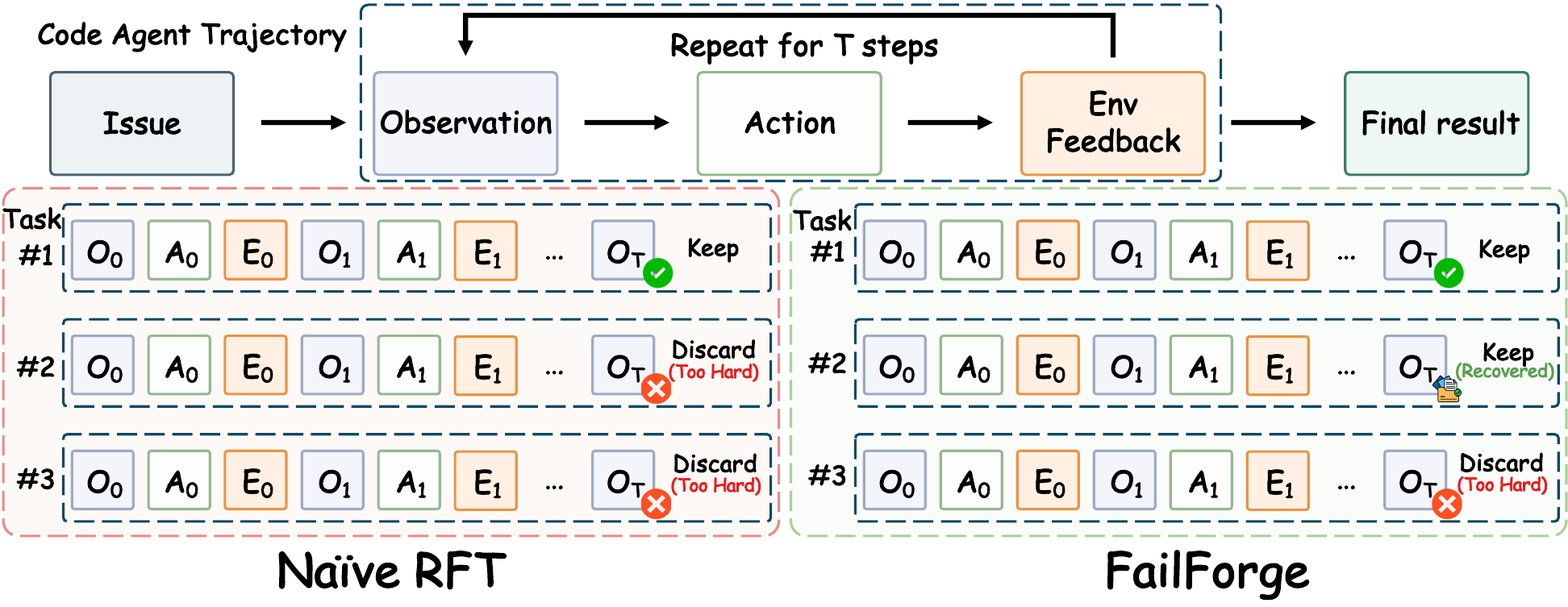}
    \caption{Recovering failed trajectories with skill guidance.}
    \label{fig:figure_in_intro}
\end{figure}
However, RFT carries a structural inefficiency that grows more severe as tasks become harder. Tasks on which every sampled trajectory fails contribute absolutely nothing to the learning process. Even frontier models fail consistently on a substantial fraction of SWE tasks, meaning the entire sampling budget spent on those tasks is wasted~\cite{yu2026dapo,wu2026learn}. Furthermore, the discarded tasks are not a random subset. They sit precisely at the capability frontier of the model, which is exactly where learning signals would be most valuable~\cite{bengio2009curriculum,jiang2021prioritized}. The remedy is not simply to sample more. Scaling the per-task budget from 5x to 10x recovers only a small portion of failed tasks. This indicates that persistent failures reflect missing capabilities rather than merely sampling variance. Figure~\ref{fig:figure_in_intro} illustrates both the structure of a code agent trajectory and the key inefficiency of standard RFT, which discards failed instances that \method{} can partially recover through skill guidance.

Existing methods that salvage discarded trajectories mostly operate within a failed trajectory. They mine useful actions from failed rollouts~\cite{lan2025exploring}, mask the loss on erroneous steps~\cite{slinko2026step}, reweight negative trajectories by their reward~\cite{liu2026rift}, or fuse incorrect and correct rollouts into one trace~\cite{deng2026beyond}, and a related line rebalances which solved instances are sampled~\cite{koh2026adastar}. This is fragile on long horizon software engineering tasks, where sparse outcome feedback makes it difficult to attribute failure to individual actions
~\cite{jimenez2024swebench,tan2026hindsight}. An agent may skip a relevant piece of code during early exploration and, lacking that context, later edit in the wrong order, so the patch fails even though no individual step is clearly incorrect. When guidance is derived at the step or fragment level, such failures leave little to reuse and these methods reduce to standard RFT~\cite{zelikman2022star}. What they extract is also tied to the specific instance rather than a transferable strategy, and most are studied on mathematical reasoning rather than repository tasks, where longer contexts make step level judgments less reliable.

Our key observation is that many persistent failures stem not from missing knowledge about the instance, but from the absence of transferable procedural competence, the reusable strategy for how to investigate a problem. This includes knowing how to localize a fault across a large repository, how to read a complex test harness, and how to structure an efficient investigation. Motivated by this observation, we introduce \method{}, a pipeline that induces such a skill from the failed trajectories of an instance and re-attempts the instance with the skill supplied as guidance. Unlike guidance that points to the location of a bug, the skill states a general strategy that narrows the search over investigation paths without revealing the answer. Under the same sampling budget, this recovers nearly twice as many solvable instances as additional rounds of standard sampling.

Our main contributions are summarized as follows:
\begin{itemize}
\item We introduce \method{}, which turns failed SWE instances discarded by standard RFT into transferable procedural supervision. It diagnoses failures, uses the resulting leakage-filtered skill to synthesize a successful trajectory, and removes the skill before training so that the procedure is internalized without inference-time guidance.

\item We demonstrate that \method{} produces stronger supervision than additional sampling or instance level hints. On SWE-bench Verified, it improves pass@1 by $6.6$ points over a strong RFT baseline, with all gains coming from instances discarded by standard RFT.

\item We show that the recovered supervision transfers beyond the source instance, improves the value of each training sample, and induces more systematic problem solving behavior across diverse tasks.
\end{itemize}

\section{Related Work}
\subsection{Code Agents for Software Engineering}
LLM-based agents have become a standard paradigm for repository-level software engineering, with systems such as SWE-agent, OpenHands, and Agentless combining repository navigation, code editing, and execution-based verification~\cite{yang2024sweagent,wang2025openhands,xia2024agentless}. Benchmarks and executable training environments including SWE-bench, SWE-Gym, SWE-smith, and R2E-Gym enable large-scale evaluation and trajectory collection~\cite{jimenez2024swebench,pan2024training,yang2026swe,jain2025r2e}. Most existing training pipelines retain verified trajectories while discarding failed ones. \method{} instead targets instances for which repeated agent rollouts all fail and converts them into additional supervision.

\subsection{Rejection Sampling Fine-Tuning}
RFT, or filtered behavior cloning, trains a model on sampled trajectories that pass an outcome verifier~\cite{yuan2023scaling,gulcehre2023reinforced}, and STaR applies this self-bootstrapping principle to reasoning~\cite{zelikman2022star}. A related zero-signal problem occurs in group-based RL, where uniformly failed groups yield no useful gradient and dynamic-sampling methods such as DAPO filter these groups rather than recover supervision from them~\cite{shao2024deepseekmath,yu2026dapo}. Other work reuses failures by extracting useful actions, masking erroneous steps, reweighting negative trajectories, or combining failed and successful traces~\cite{lan2025exploring,slinko2026step,liu2026rift,deng2026beyond}. These approaches mainly operate on steps or fragments within existing trajectories, which is difficult for repository-level tasks because sparse outcome feedback and long-range dependencies make individual failure steps hard to identify. \method{} instead diagnoses the complete failure context and uses the induced procedure to synthesize a new verified trajectory.

\subsection{Skill Internalization}
A growing line of work equips agents with reusable skills or experiential memory. Voyager, ExpeL, Agent Workflow Memory, and ReasoningBank extract strategies or workflows and retrieve them as additional context at inference time~\cite{wang2023voyager,zhao2024expel,wang2024agent,ouyang2025reasoningbank}, while reflection and self-correction methods likewise depend on iterative inference-time feedback or stored reflections~\cite{shinn2023reflexion,madaan2023self,gou2024critic}. Hindsight Hint Distillation derives an instance-level hint from failed self-rollouts, uses it to scaffold a successful SWE trajectory, and distills the result into a hint-free policy~\cite{wang2026hindsight}. In contrast, \method{} diagnoses full failed trajectories, test outputs, and reasoning traces to induce a high-level, non-task-specific procedural skill. It explicitly filters leakage and removes the injected skill before training, so that the recovered supervision targets transferable investigation competence rather than guidance bound to the source instance.

\section{Methodology}
\begin{figure*}
    \centering
    \includegraphics[width=\linewidth]{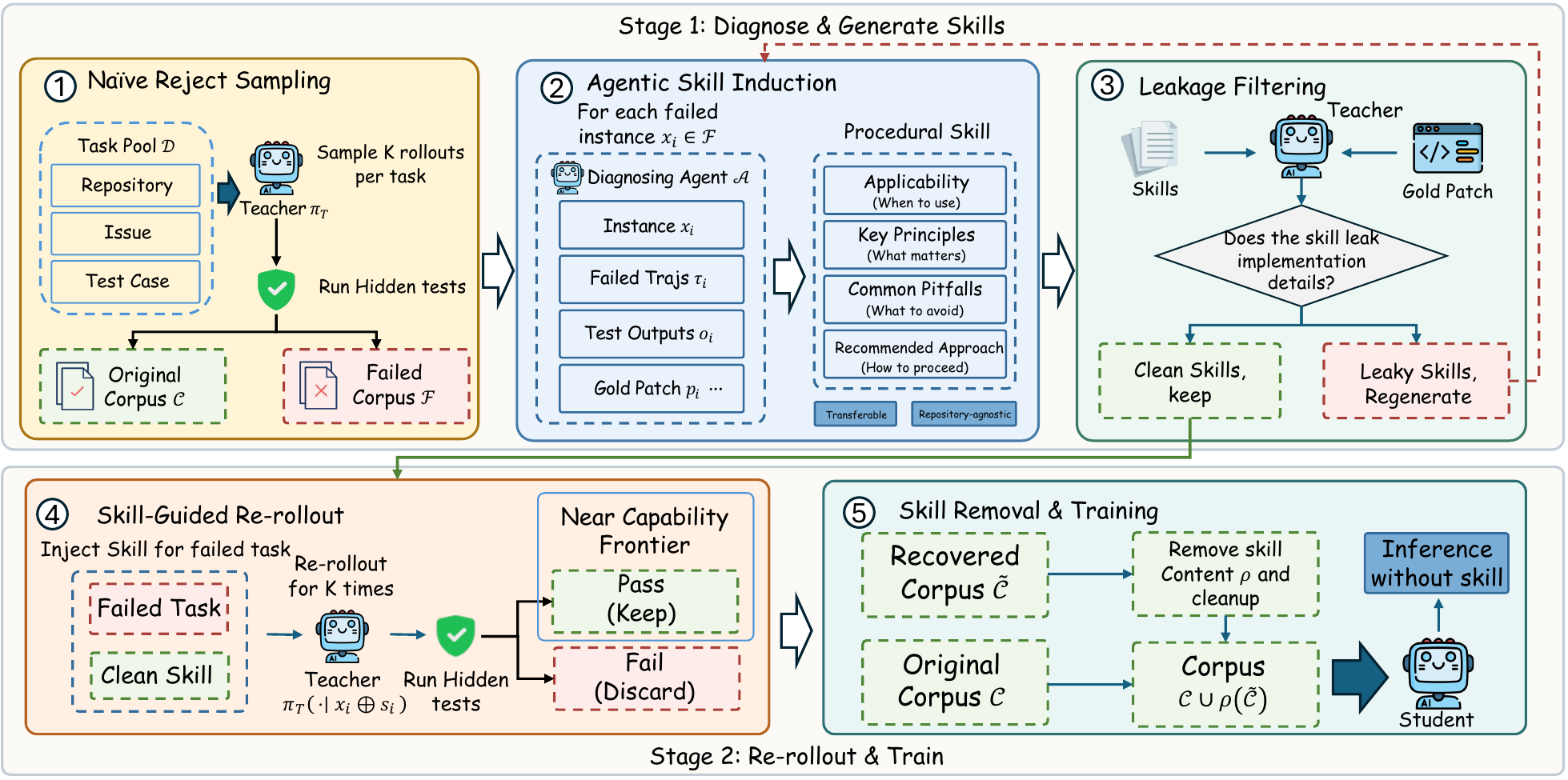}
    \caption{\textbf{Overview of \method{}.}
FailForge diagnoses persistently failed instances, induces leakage-filtered procedural skills, uses them to guide new rollouts, and removes the skills before fine-tuning.}
    \label{fig:overview}
\end{figure*}

\subsection{Problem Definition}
We train a code agent by RFT on verifiable
software engineering tasks. We adopt the distillation regime, in which
trajectories are drawn from a teacher policy $\pi_{\mathrm{T}}$ and a distinct student $\pi_\theta$ is fine-tuned on those that pass verification. Let $\mathcal{D} = \{x_i\}_{i=1}^{M}$ be the task pool, where each $x_i$ pairs a repository snapshot and a natural-language issue with a hidden test suite $T_i$ that acts as an execution-based verifier. A rollout of $\pi_{\mathrm{T}}$ on $x_i$ yields a trajectory\footnote{A trajectory of the harness including tool calls, edits, and observations.} $\tau$ and a final patch, and the verifier returns a binary reward $R(x_i, \tau) \in \{0, 1\}$, with $R = 1$ if the patch passes all tests in $T_i$.

Standard RFT samples $K$ trajectories per task and retains only the successful pairs, $\mathcal{C} = \{(x_i, \tau) \mid R(x_i, \tau) = 1\}$, and discards every task whose $K$ rollouts all fail. The persistently failed set is:
\[
\mathcal{F} = \{\, x_i \in \mathcal{D} \;\mid\; R(x_i, \tau_{i}^{(k)}) = 0
\ \ \forall k \in \{1,\dots,K\} \,\}.
\]
These persistently failed instances sit at the teacher's capability frontier, offering no trajectory to imitate yet the most valuable signal to learn from. Our goal is to recover trainable supervision from $\mathcal{F}$ for an unconditional student that samples from $\pi_\theta(\cdot \mid x_i)$ with no auxiliary context at test time. This sets it apart from methods that scaffold the policy at inference time, which condition on retrieved skills, stored reflections, or prepended hints and thus inherit the retrieval limits of their guidance channel. The difficulty is that supervision must be manufactured from failures without leaking the manufacturing signal into the policy's test time interface.

\subsection{Method}\label{sec:method}
A natural attempt to rescue $\mathcal{F}$ is to simply sample more trajectories per task, but increasing the per-task budget recovers only a small fraction of $\mathcal{F}$, and a taxonomy of the residual failures shows that the missing ingredient is rarely knowledge about the specific instance. What the agent lacks instead is transferable procedural competence, the kind of methodology that carries over from one repository to the next. An instance level hint, bound to a single bug, cannot convey such methodology, which is precisely why supplying more hints does not close the gap. Concretely, a competent agent treats the existing test suite as the specification of the intended interface, audits every site that produces or consumes a changed identifier rather than patching only the one that surfaced the error, and verifies that a new helper is importable where the tests expect it and valid for the downstream systems that consume its output. \method{} supplies this transferable competence as an explicit synthesis time scaffold and then distills it away, through a recovery pipeline over $\mathcal{F}$ comprising five stages, see Figure~\ref{fig:overview}.

\paragraph{Naive RFT and failure identification.}

We first run standard RFT. For each task $x_i \in \mathcal{D}$ we sample $K$ teacher trajectories $\{\tau_i^{(k)}\}_{k=1}^{K} \sim \pi_{\mathrm{T}}(\cdot \mid x_i)$ and verify each against $T_i$, yielding the initial corpus $\mathcal{C}$ of successful pairs and the persistently failed set $\mathcal{F}$. The subsequent four stages will operate exclusively on $\mathcal{F}$.
 
\paragraph{Agentic skill induction.}
For each failed instance $x_i \in \mathcal{F}$, we invoke a diagnosing agent
$\mathcal{A}$ that produces a skill:
\[
s_i = \mathcal{A}\!\left(x_i,\, \{\tau_i^{(k)},\,o_i^{(k)}\}_{k=1}^{K},\, p_i\right),
\]
where $\tau_i^{(k)}$ are the failed trajectories, $o_i^{(k)}$ their test outputs and reasoning traces, and $p_i$ the gold patch provided as reference material. This assumption is practical for offline SWE data construction because training instances are commonly derived from historical issue fix pairs, for which the merged patch is naturally available. The agent reasons over the full set of failed trajectories, test outputs, and reasoning traces to isolate the root cause.

It emits a structured artifact rather than free form
text, following a fixed schema of four parts, an \emph{applicability condition} that states when the skill is relevant, a set of \emph{key principles} that name the invariants a correct fix must respect, the \emph{common pitfalls} that may lead to similar failures, and a \emph{recommended approach} that gives an ordered investigation and repair procedure. Every part is phrased at the level of methodology, naming the concepts a fix depends on, such as cross module identifier consistency or module import visibility, without referring to the specific repository or bug. The resulting $s_i$ is thus a concise, high level, and \emph{non-task-specific} procedural guideline. Because $s_i$ supplies a strategic prior rather than the solution itself, it narrows the agent's search over investigation strategies without revealing where the bug is, which is what enables the guided re-attempt to succeed where unguided sampling repeatedly failed.

\paragraph{Leakage filtering.}

The gold patch is used only as privileged supervision during offline synthesis on training instances. The training repositories are disjoint from those in both evaluation benchmarks, and neither the gold patch nor the induced skill is available at test time. Therefore, the privileged supervision does not expose solutions from the evaluation instances. It may nevertheless introduce solution specific information into the induced skill, making a training instance easier to recover while weakening the interpretation of the skill as a transferable procedure.

To reduce this risk, we pass each pair $(s_i,p_i)$ to a separate judge model $\mathcal{J}$ defined as:
\[
    \mathcal{J}(s_i, p_i) =
    \begin{cases}
    1 & \text{if } s_i \text{ leaks implementation detail of } p_i,\\
    0 & \text{otherwise,}
    \end{cases}
\]
applying a concrete reconstruction test, under which $s_i$ leaks if a
reader could reconstruct $p_i$ from it, whether by copying identifiers, mirroring the patch's control-flow structure even under renaming, or prescribing instance-specific fix steps. High level methodology that merely names a relevant concept is considered clean. 

A skill with $\mathcal{J}(s_i,p_i)=1$ is regenerated until it passes the filter or the retry budget is exhausted. We retain only skills judged clean by $\mathcal{J}$. We calibrate the judge against two independent human annotators. On the items for which the annotators agree, the judge matches the human consensus on $96.7\%$ of the samples, with Cohen's $\kappa=0.805$~\cite{cohen1960coefficient}. Treating leakage as the positive class, it achieves $0.875$ precision and $0.778$ recall. These results support GPT-5.4 as a scalable leakage screening mechanism. Calibration results are provided in Supplementary Section~B.2.

\paragraph{Skill-guided re-rollout.}
We re-attempt each failed instance with the retained skill $s_i$ injected into the teacher's context. Writing $\pi_{\mathrm{T}}(\cdot \mid x_i \oplus s_i)$ for the teacher conditioned on the task augmented with the skill ($\oplus$ denotes context injection, e.g., writing $s_i$ into \texttt{CLAUDE.md} for Claude Code or into the system prompt for OpenHands), we sample guided trajectories $\tilde{\tau}_i \sim \pi_{\mathrm{T}}(\cdot \mid x_i \oplus s_i)$ and keep those that pass verification, forming the recovered corpus:
\[
\tilde{\mathcal{C}} = \{\, (x_i, \tilde{\tau}_i) \mid x_i \in \mathcal{F},\,
R(x_i, \tilde{\tau}_i) = 1 \,\}.
\]
These recovered trajectories constitute additional supervision from
instances that standard RFT discards entirely.

\paragraph{Skill removal and training.}
The skill enters only at synthesis time, as part of the teacher's context. We treat $s_i$ as a latent guidance variable that steers the teacher into a high reward region of trajectory space, and the removal operator $\rho$ then strips it out before training. Concretely, $\rho$ removes the injected skill itself from the recovered trajectory but leaves the agent's own actions and reasoning fully untouched, so that the student is trained to reproduce $\rho(\tilde{\tau}_i)$ from $x_i$
alone. We then fine-tune $\pi_\theta$ on the union of the original and decontaminated recovered corpora by minimizing the standard negative log-likelihood objective:
\[
\mathcal{L}(\theta) = -\!\!\!\sum_{(x_i, \tau) \,\in\, \mathcal{C}}\!\!\!
\log \pi_\theta(\tau \mid x_i)
\;-\!\!\!\sum_{(x_i, \tilde{\tau}_i) \,\in\, \tilde{\mathcal{C}}}\!\!\!
\log \pi_\theta\!\left(\rho(\tilde{\tau}_i) \mid x_i\right).
\]
Training thus distills the teacher's skill-conditioned success,
$\pi_{\mathrm{T}}(\cdot \mid x_i \oplus s_i)$, into a skill-free student, $\pi_\theta(\cdot \mid x_i)$. Because the target retains no trace of $s_i$, reproducing the recovered behavior requires encoding the underlying procedure in the weights rather than copying a visible hint.
\section{Experiments}
\subsection{Experimental Setup}

\paragraph{Datasets and evaluation.}
We build our training data from SWE-Gym~\cite{pan2024training}, which provides $2{,}401$ Python task instances\footnote{We discard 37 tasks without docker images.}, each paired with an executable environment and a hidden test suite. We evaluate on SWE-bench Verified ~\cite{jimenez2024swebench}, the $500$-instance human-validated subset that is the standard benchmark for execution-based issue resolution, and on SWE-bench MultiLingual~\cite{yang2026swe} which spans multiple languages to examine out-of-distribution transfer. The repositories used for training in SWE-Gym have no overlap with those in either SWE-bench Verified or SWE-bench MultiLingual, ensuring repository-level separation between training and evaluation. Following common SWE-bench evaluation practice, we allow up to five independent attempts per instance and stop once a successful patch is found. We report resolve rate as \emph{pass@k} for $k\in\{1,3,5\}$, where pass@$k$ measures the fraction of instances solved within the first $k$ attempts.

\paragraph{Agent harness and models.}
All trajectories are generated by a single teacher policy $\pi_{\mathrm{T}}$ built on Kimi-K2.6~\cite{team2026kimi} and collected with the OpenHands agent scaffold~\cite{wang2025openhands}. The same teacher also performs skill induction and the guided re-rollouts, while GPT-5.4 is used as the leakage filter. The student policy $\pi_\theta$ that we fine-tune and evaluate is Qwen3.5-4B and Qwen3.5-9B~\cite{qwen35blog}. Because every method distills from the same teacher under an identical harness, any difference between methods is attributable solely to how each treats the persistently failed set $\mathcal{F}$. Full implementation and training details are provided in Supplementary Section~A.

\paragraph{Initial RFT and the failed set.}
We run standard RFT by sampling $K = 5$ trajectories per task from the teacher policy $\pi_{\mathrm{T}}$ and verifying each against its test suite. Of the $2{,}401$ tasks, $1{,}569$ are solved by at least one of the five rollouts and their successful trajectories form the RFT corpus $\mathcal{C}$. The remaining $832$ tasks, on which all five rollouts fail, constitute the persistently failed set $\mathcal{F}$ that our recovery pipeline targets.

\subsection{Baselines}
We compare \method{} against the following baselines, all trained from the same base model and initial RFT corpus $\mathcal{C}$. For every instance in $\mathcal{F}$, RFT + More Sampling, RFT + Hint, and \method{} each use exactly five additional teacher rollouts. To ensure a fair comparison, we retain only the first successful trajectory among these additional rollouts and add it to the training corpus. They differ only in whether these rollouts are unguided, conditioned on an instance-level hint, or conditioned on a procedural skill.

\paragraph{Base.} The base model without any fine-tuning.

\paragraph{RFT.} Standard rejection sampling fine-tuning on the successful trajectories $\mathcal{C}$ from the initial $K = 5$ sampling. All methods below add data on top of the same $\mathcal{C}$.

\paragraph{RFT + More Sampling.}
To test whether persistent failures are merely a matter of sampling variance, we draw five additional unguided teacher trajectories for every instance in $\mathcal{F}$, doubling the total per-task budget from $5$ to $10$. Successful trajectories are added to $\mathcal{C}$ for training.

\paragraph{RFT + Hint.}
Following HHD~\cite{wang2026hindsight}, we generate a short instance level hint from the failed trajectories and gold patch. The hint guides five additional rollouts, after which the first successful trajectory is retained and the hint is removed before training. This baseline matches \method{} in rollout budget and training procedure, differing only in the form of guidance.

\subsection{Main Results}
\newcommand{\best}[1]{\cellcolor{cyan!15}\textbf{#1}}
\newcommand{\second}[1]{\cellcolor{cyan!6}#1}

\begin{table*}[t]
\centering
\begin{tabular}{ll ccc | ccc}
\toprule
\rowcolor{gray94}
 & & \multicolumn{3}{c|}{\textbf{Qwen3.5-4B}} & \multicolumn{3}{c}{\textbf{Qwen3.5-9B}} \\
\rowcolor{gray94}
\textbf{Benchmark} & \textbf{Method} & \textbf{pass@1} & \textbf{pass@3} & \textbf{pass@5} & \textbf{pass@1} & \textbf{pass@3} & \textbf{pass@5} \\
\midrule
\multirow{5}{*}{\textbf{SWE-Bench Verified}}
  & Base                               & 54.0 & 68.8 & 72.0 & 61.2 & 71.4 & 74.6 \\
  & RFT (5$\times$)                    & 59.6 & 70.4 & 73.0 & 63.6 & \second{73.8} & 75.4 \\
  & RFT (10$\times$)                   & 63.0 & \second{71.6} & \second{75.6} & \second{66.4} & 72.4 & \second{75.6} \\
  & RFT (5$\times$) + Hint (5$\times$) & \second{63.8} & 71.4 & 73.4 & 65.2 & 73.0 & 74.4 \\
  & \method{} (Ours)                   & \best{66.2} & \best{73.4} & \best{76.0} & \best{67.2} & \best{74.0} & \best{76.0} \\
\midrule
\multirow{5}{*}{\textbf{SWE-Bench MultiLingual}}
  & Base                               & 35.7 & 53.0 & 56.3 & 47.0 & 62.0 & 65.0 \\
  & RFT (5$\times$)                    & 52.3 & 62.3 & 65.3 & 59.7 & 66.7 & 70.7 \\
  & RFT (10$\times$)                   & \best{55.7} & \best{67.0} & \second{70.0} & \best{61.3} & 69.0 & \second{72.7} \\
  & RFT (5$\times$) + Hint (5$\times$) & 51.7 & 64.3 & 67.3 & 60.7 & \second{69.3} & 71.3 \\
  & \method{} (Ours)                   & \best{55.7} & \best{67.0} & \best{70.7} & \best{61.3} & \best{71.0} & \best{74.0} \\
\bottomrule
\end{tabular}
\caption{Resolve rate (\%) on SWE-bench Verified and SWE-bench MultiLingual with OpenHands harness.}
\label{tab:main}
\end{table*}

Table~\ref{tab:main} reports the results under the OpenHands harness, since all methods share the same initial corpus $\mathcal{C}$, every difference is attributable to how each treats the discarded set $\mathcal{F}$. \method{} attains $66.2$ pass@1, improving the strong RFT baseline by $+6.6$ points, with consistent gains of $+3.0$ at pass@3 and $+3.0$ at pass@5, indicating that the recovered trajectories sharpen single-attempt solving rather than merely widening coverage across $k$.

The two strongest alternatives fall short for complementary reasons. Doubling the rollout budget reaches only $63.0$ pass@1, confirming that persistent failures cannot be addressed by additional sampling alone. Instance-level hints recover more trajectories than \method{}, yet produce a weaker student, reaching only $63.8$ pass@1 compared with $66.2$ for \method{} and trailing it at every $k$. Recovery count is therefore not a sufficient measure of supervision quality. Hints make individual instances easier to solve, whereas procedural skills yield recovered trajectories with greater downstream training value.

The improvement also persists across model scales and evaluation domains. On SWE-bench Verified, \method{} raises the 9B model from $63.6$ to $67.2$ pass@1 and achieves the best result at every $k$. On SWE-bench MultiLingual, it improves the 4B RFT baseline from $52.3$ to $55.7$ pass@1 and the 9B baseline from $59.7$ to $61.3$. Although additional sampling matches \method{} at pass@1 in some MultiLingual settings, \method{} performs better at larger $k$, reaching $70.7$ and $74.0$ pass@5 for the 4B and 9B models. These results suggest that the recovered supervision is not specific to one student scale or to the Python distribution used for training.

\paragraph{Cost efficiency.}
\begin{table}[t]
\centering
\small
\setlength{\tabcolsep}{4pt}
\begin{tabular}{l c c c c c}
\toprule
\rowcolor{gray94}
\textbf{Method}
& \textbf{$N_{\text{rec}}$}
& \textbf{Rec.\%}
& \textbf{Tok/rec}
& \textbf{pass@1}
& \textbf{Tok/$\Delta$} \\
\rowcolor{gray94}
& & & \textbf{(M)} & & \textbf{(B)} \\
\midrule
RFT (5$\times$)  & --  & --   & --    & 59.6 & --   \\
RFT (10$\times$) & 119 & 14.3 & 123.3 & 63.0 & 4.32 \\
RFT + Hint       & 352 & 42.3 & 35.9  & 63.8 & 3.01 \\
\rowcolor{gray94}
\method{}        & 218 & 26.2 & 64.1  & \textbf{66.2} & \textbf{2.12} \\
\bottomrule
\end{tabular}
\caption{\textbf{Cost and performance on the persistently failed set
$\mathcal{F}$.}
The three recovery methods use comparable teacher token budgets on
$\mathcal{F}$. $N_{\text{rec}}$ denotes the number of recovered instances, Rec.\% denotes the recovery rate over $\mathcal{F}$, Tok/rec denotes the teacher tokens per recovered instance in millions, and Tok/$\Delta$ denotes the teacher tokens per point of pass@1 improvement over RFT (5$\times$) in billions.}
\label{tab:cost}
\end{table}

As shown in Table~\ref{tab:cost}, instance level hints recover more instances and require fewer tokens for each recovery, whereas \method{} incurs additional costs for skill induction and leakage filtering. However, recovery count alone does not determine the training value of the resulting trajectories. \method{} achieves the lowest token cost for each point of pass@1 improvement, requiring $2.12$B tokens per point, compared with $3.01$B for instance level hints and $4.32$B for additional sampling. The additional induction cost is therefore offset by the greater downstream value of the recovered trajectories.

\paragraph{Recovered data quality.}
\begin{table}[t]
\centering
\begin{tabular}{l ccc}
\toprule
\rowcolor{gray94}
\textbf{Fill data} & \textbf{pass@1} & \textbf{pass@3} & \textbf{pass@5} \\
\midrule
Original         & 59.6 & 70.4 & 73.0 \\
\rowcolor{gray94}
\method{}        & \textbf{63.8} & \textbf{73.2} & \textbf{75.0} \\
\bottomrule
\end{tabular}
\caption{\textbf{Data quality at a fixed training budget.} A fixed-size portion of
the original RFT corpus is replaced with an equal number of recovered
trajectories, holding the total training set size constant.}
\label{tab:dataquality}
\end{table}

To verify the gain is not merely a data volume effect, we hold the total number of training trajectories fixed and replace a fixed-size portion of the original RFT corpus $\mathcal{C}$ with an equal number of \method{} recovered trajectories, see Table~\ref{tab:dataquality}. Because the training set size is identical, any difference isolates per-sample quality. Swapping in the recovered trajectories raises pass@1 from $59.6$ to $63.8$, showing that a recovered trajectory carries more useful supervision per sample than the original RFT trajectory it displaces.

\paragraph{Cross harness transfer.}

\begin{table*}[t]
\centering
\begin{tabular}{ll ccc | ccc}
\toprule
\rowcolor{gray94}
 & & \multicolumn{3}{c|}{\textbf{Qwen3.5-4B}} & \multicolumn{3}{c}{\textbf{Qwen3.5-9B}} \\
\rowcolor{gray94}
\textbf{Benchmark} & \textbf{Method} & \textbf{pass@1} & \textbf{pass@3} & \textbf{pass@5} & \textbf{pass@1} & \textbf{pass@3} & \textbf{pass@5} \\
\midrule
\multirow{5}{*}{\textbf{SWE-Bench Verified}}
  & Base                               & 47.6 & 65.0 & 69.0 & 60.6 & 75.0 & 77.2   \\
  & RFT (5$\times$)                    & 59.6 & \best{75.0} & 78.0 & 61.0 & \second{75.2} & 77.2 \\
  & RFT (10$\times$)                   & 59.4 & 72.2 & 77.0 & 61.4 & 75.0 & \second{78.2} \\
  & RFT (5$\times$) + Hint (5$\times$) & \second{60.0} & \second{74.2} & \second{78.2} & \second{62.2} & 74.6 & 77.4 \\
  & \method{} (Ours)                   & \best{62.6} & 74.0 & \best{78.4} & \best{63.0} & \best{75.8} & \best{78.6}  \\
\midrule
\multirow{5}{*}{\textbf{SWE-Bench MultiLingual}}
  & Base                               & 39.0 & 54.7 & 59.3 & 48.0 & \best{68.0} & \best{74.7} \\
  & RFT (5$\times$)                    & 41.0 & 60.3 & 65.3 & 49.7 & 66.0 & 69.3 \\
  & RFT (10$\times$)                   & \second{47.7} & 62.0 & 65.7 & 48.0 & 66.7 & 70.3 \\
  & RFT (5$\times$) + Hint (5$\times$) & 45.0 & \second{63.3} & \second{68.0} & \second{50.7} & \second{67.0} & 71.0 \\
  & \method{} (Ours)                   & \best{50.3} & \best{65.3} & \best{68.3} & \best{51.0} & 66.0 & \second{71.3} \\
\bottomrule
\end{tabular}
\caption{Resolve rate (\%) on SWE-bench Verified and SWE-bench MultiLingual with Claude Code harness.}
\label{tab:main2}
\end{table*}

So far all trajectories, including the recovered ones, are generated with the OpenHands harness. Table~\ref{tab:main2} tests whether the learned behavior carries to a different harness by evaluating the models under Claude Code. A model trained purely on OpenHands trajectories overfits to that interface and, when evaluated under Claude Code, hallucinates OpenHands style tool calls the new harness does not expose. To prevent this we add $100$ trajectories sampled with Claude Code to every method's training mixture, while all recovered supervision and induced skills remain OpenHands only, so what we measure is transfer of behavior rather than tool format. \method{} again attains the best pass@1 on both benchmarks, $62.6$ on Verified and $50.3$ on MultiLingual, indicating the recovered investigation procedure is not tied to the harness it was collected under.

\subsection{Ablation Study}

We ablate the three pipeline components of \method{}, namely skill-guided re-rollout (RR), leakage filtering (LF), and skill removal before training (SR), in Table~\ref{tab:ablation}, and the two skill-synthesis design choices in Table~\ref{tab:design}.

\begin{table}[t]
\centering
\begin{tabular}{ccc | ccc}
\toprule
\rowcolor{gray94}
\textbf{LF} & \textbf{RR} & \textbf{SR} & \textbf{pass@1} & \textbf{pass@3} & \textbf{pass@5} \\
\midrule
 &  &            & 59.6 & 70.4 & 73.0 \\
\checkmark & \checkmark &            & 64.6 & 73.0 & 76.6 \\
           & \checkmark & \checkmark & 63.6 & 72.0 & 75.8 \\
\midrule
\rowcolor{gray94}
\checkmark & \checkmark & \checkmark & \textbf{66.2} & \textbf{73.4} & \textbf{76.0} \\
\bottomrule
\end{tabular}
\caption{\textbf{Component ablation on SWE-bench Verified.} LF: leakage
filtering; RR: skill-guided re-rollout; SR: skill removal before training.}
\label{tab:ablation}
\end{table}

The top row of Table~\ref{tab:ablation} disables all three components and reduces to standard RFT~$5\times$ at $59.6$ pass@1, while the full pipeline reaches $66.2$ ($+6.6$) without touching the base corpus $\mathcal{C}$, so the entire gain comes from supervision recovered from $\mathcal{F}$. Retaining the skill in the training
target drops pass@1 to $64.6$, as leaving it in context lets the student lean on the supplied strategy rather than acquire it, whereas stripping it forces the behavior into the weights. Leakage filtering acts earlier, during skill induction. Without it the induced skills drift toward answers tied to the specific instance, so the trajectories they recover follow guidance that nearly encodes the solution rather than transferable methodology, and distilling such solution-bound supervision lowers pass@1 to $63.6$, close to the hint baseline.

\begin{table}[t]
\centering
\begin{tabular}{l ccc}
\toprule
\rowcolor{gray94}
\textbf{Variant} & \textbf{pass@1} & \textbf{pass@3} & \textbf{pass@5} \\
\midrule
\method{} (full)      & \textbf{66.2} & \textbf{73.4} & \textbf{76.0} \\
\quad w/o gold patch  & 64.2 & 72.0 & 74.6 \\
\quad + trace rewrite & 62.0 & 71.0 & 74.2 \\
\bottomrule
\end{tabular}
\caption{\textbf{Skill-synthesis design choices on SWE-bench Verified.}
\emph{w/o gold patch}: the gold patch is withheld from the diagnosing agent.
\emph{trace rewrite}: recovered reasoning traces are rewritten to launder their
references to the injected skill.}
\label{tab:design}
\end{table}
Table~\ref{tab:design} ablates how skills are induced. Withholding the gold patch from the diagnosing agent drops pass@1 from $66.2$ to $64.2$. The reference solution provides useful diagnostic context. The calibrated leakage filter allows us to exploit this signal while substantially reducing the risk that solution-specific implementation details enter the induced skill. Because the skill is injected into the agent context, the recovered reasoning may explicitly refer to the supplied guidance. Rewriting each reasoning trace so the trace reads as self-derived, drops pass@1 further to $62.0$. This is counter-intuitive, since the skill text is already stripped and dangling references might be expected to read as hallucination. In practice the recovered reasoning is most useful left intact, and aggressive rewriting distorts the very investigation behavior we want the student to imitate. We therefore keep the recovered traces verbatim in our main runs.

\subsection{Cross Instance Transferability}
\label{sec:transfer}

\begin{figure}
    \centering
    \includegraphics[width=1.0\linewidth]{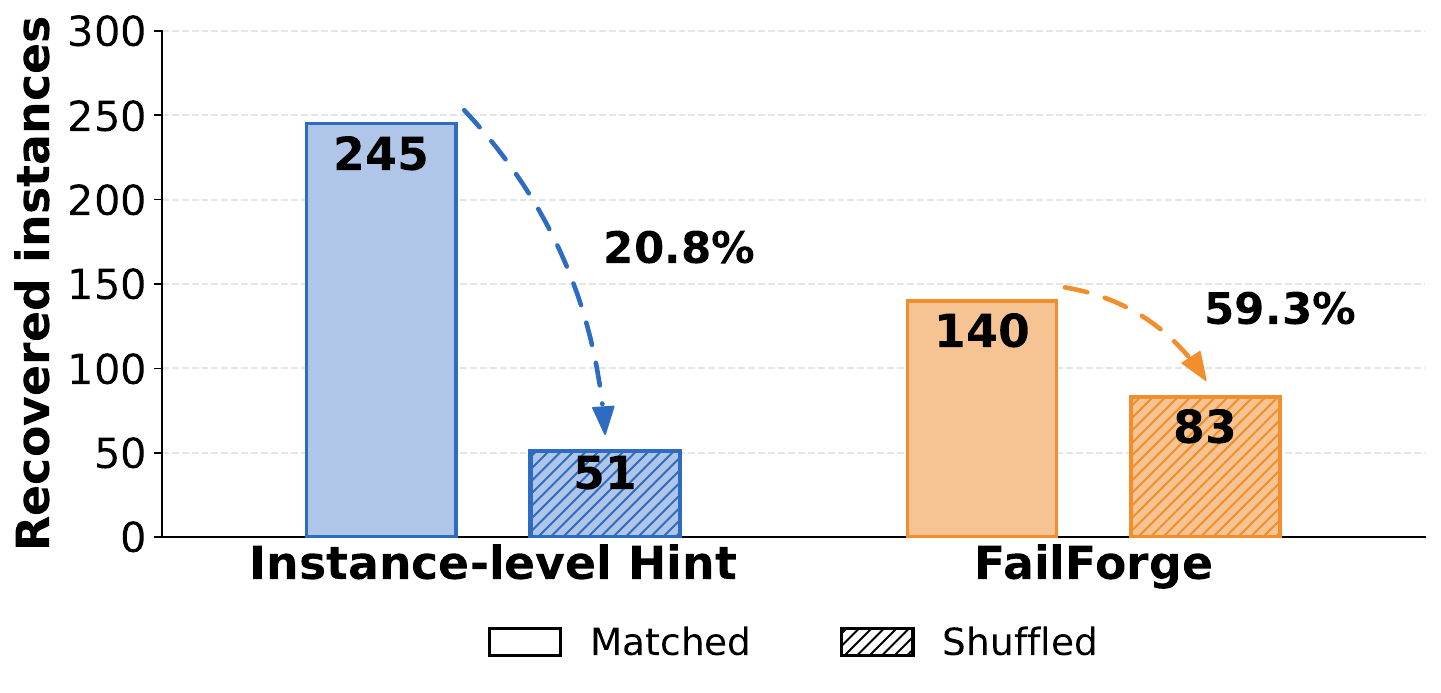}
    \caption{\textbf{Within-repository guidance transfer on $\mathcal{F}'$.}
Number of instances recovered under a single guided rollout when the guidance is
\emph{matched} to the target instance or \emph{shuffled} to a sibling instance
from the same repository.}
    \label{fig:transfer}
\end{figure}

We also test whether induced skills encode procedures that transfer beyond their source instances. From $\mathcal{F}$, we select the subset $\mathcal{F}'$ whose repositories contain at least two failed instances. Each instance receives one guided rollout under four conditions: its own skill, its own hint, a skill from a sibling instance in the same repository, or a hint from that sibling. The instance set and rollout budget are identical across conditions, so the only difference is the type and source of guidance. As shown in Figure~\ref{fig:transfer}, matched hints recover more instances than matched skills because they are tailored to the target problem. After shuffling, however, skills recover more instances and retain $59.3\%$ of their matched recovery, compared with only $20.8\%$ for hints. This comparison is conservative because the shuffled guidance still comes from the same repository and may therefore share code structure and conventions with the target instance. The substantially higher retention of skills indicates that they capture reusable investigation procedures rather than information tied to a particular solution.
\subsection{Agent Behavioral Analysis}\label{sec:behavior}

\begin{table}[t]
\centering
\setlength{\tabcolsep}{4.5pt}
\begin{tabular}{l c cc cc}
\toprule
\rowcolor{gray94}
 & & \multicolumn{2}{c}{\textbf{Loc. prec.}} & & \\
\rowcolor{gray94}
\textbf{Method} & \textbf{Resolve} & \textbf{pass} & \textbf{fail} & \textbf{Repro} & \textbf{Edits} \\
\midrule
Base                  & 54.0 & 0.46 & 0.38 & 76.2 & 16.5 \\
RFT (5$\times$)       & 59.6 & 0.44 & 0.33 & 94.0 & 10.7 \\
RFT (10$\times$)      & 63.0 & 0.47 & 0.42 & 96.0 & 10.5 \\
RFT (5$\times$)+Hint  & 63.8 & 0.52 & 0.42 & 96.8 & 10.1 \\
\rowcolor{gray94}
\method{}             & \textbf{66.2} & \textbf{0.52} & \textbf{0.45} & \textbf{97.2} & \textbf{9.9} \\
\bottomrule
\end{tabular}
\caption{\textbf{Single-attempt trajectory behavior on SWE-bench Verified.} Localization precision is the fraction of edited source files that are
gold-target files, on solved (\emph{pass}) vs.\ unsolved (\emph{fail})
instances; \emph{Repro} is the percentage of trajectories that verify before
editing; \emph{Edits} is the median number of edits. \emph{Resolve} is pass@1
from Table~\ref{tab:main}.}
\label{tab:behavior}
\end{table}

We analyze the single attempt OpenHands trajectories of all five 4B checkpoints on SWE-bench Verified along three dimensions, as shown in Table~\ref{tab:behavior}. Verification discipline measures whether the agent tests before editing, editing economy measures the number of edits, and localization precision measures the fraction of edited files that match the gold targets. Plain RFT substantially improves verification discipline and reduces the number of edits, but does not improve localization precision over the base model.

\method{} exhibits the strongest overall investigation behavior. It achieves the highest localization precision on both solved and unsolved instances, reaches the highest verification rate, and makes the fewest edits. The most pronounced difference lies in localization, especially on unsolved instances, suggesting that the recovered supervision helps the agent direct an already disciplined workflow toward more relevant code. These patterns support the interpretation
that \method{} induces a more systematic investigation procedure rather than improving any single behavior in isolation.

\subsection{Taxonomy of Extracted Skills}
\begin{figure}
    \centering
    \includegraphics[width=\linewidth]{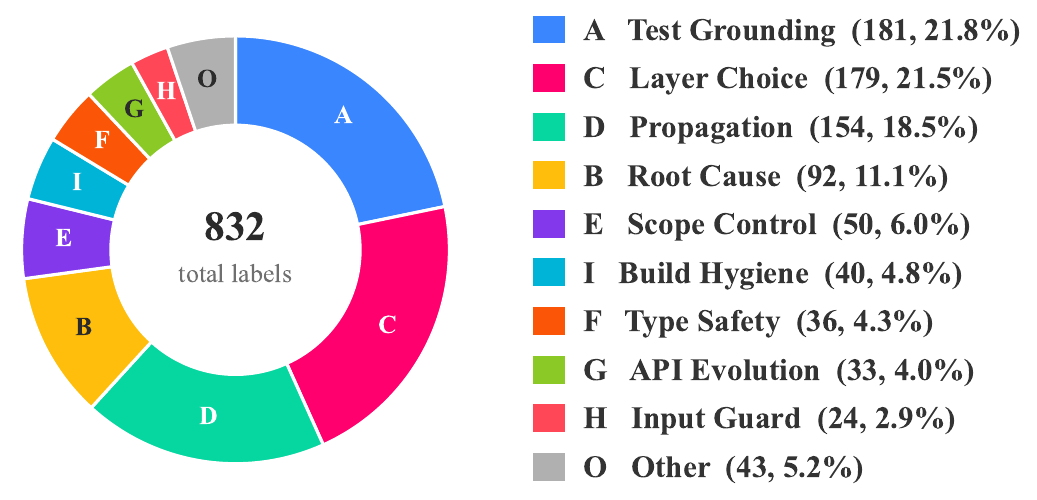}
    \caption{Distribution of induced skills across procedural capability categories.}
    \label{fig:taxonomy}
\end{figure}
We classify all $832$ induced skills into nine root-cause categories plus an \emph{Other} bucket, as shown in Figure~\ref{fig:taxonomy}. Full definitions and an inter-model agreement analysis are provided in Supplementary Section~C. On a random sample of $150$ skills, independent classifications by GLM-5.1, GPT-5.4, and Kimi-K2.6 achieve Fleiss' $\kappa=0.800$, with unanimous labels on $75\%$ of the samples. The four largest categories cover $72.8\%$ of all skills and concern test-contract grounding, fix placement, change propagation, and root-cause tracing. These recurring procedural capabilities align with the localization gap observed in Table~\ref{tab:behavior}.

\section{Conclusion}
We present \method{}, a framework that converts failed software engineering instances into training supervision by inducing transferable procedural skills for guided recovery. At a matched budget, \method{} recovers nearly twice as many instances as additional sampling, improves SWE-bench Verified by $6.6$ points over a strong RFT baseline, and induces more systematic and transferable problem-solving behaviors.

\bibliography{aaai2027}
\end{document}